# A frugal Spiking Neural Network for unsupervised classification of continuous multivariate temporal data

Sai Deepesh Pokala[1], Marie Bernert[1], Takuya Nanami[2], Takashi Kohno[2], Timothée Lévi[3], Blaise Yvert[1]

1. Univ. Grenoble Alpes, INSERM, U1216, Grenoble Institut Neurosciences, Grenoble, France
2. Institute of Industrial Science, The University of Tokyo, Meguro, Tokyo, 153-8505, Japan
3. IMS Laboratory, UMR5218, University of Bordeaux, Talence, France

## Abstract

As neural interfaces become more advanced, there has been an increase in the volume and complexity of neural data recordings. These interfaces capture rich information about neural dynamics that call for efficient, real-time processing algorithms to spontaneously extract and interpret patterns of neural dynamics. Moreover, being able to do so in a fully unsupervised manner is critical as patterns in vast streams of neural data might not be easily identifiable by the human eye. Formal Deep Neural Networks (DNNs) have come a long way in performing pattern recognition tasks for various static and sequential pattern recognition applications. However, these networks usually require large labeled datasets for training and have high power consumption preventing their future embedding in active brain implants. An alternative aimed at addressing these issues are Spiking Neural Networks (SNNs) which are neuromorphic and use more biologically plausible neurons with evolving membrane potentials. In this context, we introduce here a frugal single-layer SNN designed for fully unsupervised identification and classification of multivariate temporal patterns in continuous data with a sequential approach. We show that, with only a handful number of neurons, this strategy is efficient to recognize highly overlapping multivariate temporal patterns, first on simulated data, and then on Mel Cepstral representations of speech sounds and finally on multichannel neural data. This approach relies on several biologically inspired plasticity rules, including Spike-timing-dependent plasticity (STDP), Short-term plasticity (STP) and intrinsic plasticity (IP). These results pave the way towards highly frugal SNNs for fully unsupervised and online-compatible learning of complex multivariate temporal patterns for future embedding in dedicated very-low power hardware.

## Introduction

Neural interfacing with arrays of electrodes is key for understanding the dynamics of the CNS and for the development of neural prostheses for rehabilitation in case of severe paralysis[1–9]. The amount of data produced by recent cortical recording devices[10–14] has become enormous and their richness is difficult to access using conventional methods. In particular, whether specific spatiotemporal patterns exist in multichannel neural data might not necessarily be obvious to determine, and the corresponding patterns underlying specific neural functions difficult to apprehend. To this end, fully unsupervised approaches that could extract patterns of neural dynamics from the large flow of data produced by neural implants would bring invaluable perspective to better understand the dynamics of the brain that underly behavior and to identify non-obvious behaviorally relevant neural features. Furthermore, the long-term objective of seamlessly integrating neural processing directly into implantable devices emphasizes the critical need for algorithms that are both efficient and online-compatible, while also being low in power consumption.

Over the past decade, formal deep neural networks (DNNs) have reached unprecedented interest to learn patterns within large amounts of data[15–19]. This second generation of artificial neural networks (ANNs) is now extensively used and ubiquitous in many applications. Yet, despite their capabilities, they face two main drawbacks to envision their embedding in neural interfaces. Firstly, their reliance on mainly supervised learning techniques such as backpropagation[20] requires labeled datasets, posing a significant obstacle in applications where such labeled data is either limited, or difficult to obtain as in the case of large-scale neural recordings. Secondly, the computational demands of these networks often call for the use of specialized hardware like GPUs or TPUs to optimize their numerous parameters that need to be maintained in memory and learned through the minimization of a global loss function. Therefore, DNNs are unlikely to be pertinent candidates to eventually embed automated neural processing algorithms into future intelligent neural implants for real-time identification and extraction of complex features from large scale neural recordings.

By contrast, spiking neural networks (SNNs) are neuromorphic ANNs that model the membrane potential of their neural elements[21–23]. They typically rely on smaller numbers of parameters, and integrate biomimetic plasticity rules found in living neural networks such as spike-timing-dependent plasticity (STDP)[24,25] or other post-synaptic rules. This third generation of ANNs is thus radically different from DNNs, as learning becomes local at each synapse and neural element based on the dynamics of pre- and post-synaptic neurons. This removes the necessity for large global memory storage and energy-consuming global minimization. Based on their local learning rules, SNNs can self-configure in a fully unsupervised way solely based on their inputs to automatically recognize patterns hidden in the data[26–28], and while they are also capable of supervised learning through surrogate gradient backpropogation[29–32], they require less data for training than traditional DNNs[33–35]. Finally, SNNs are compatible with very-low power neuromorphic hardware emulating spiking neurons[36–40] and novel materials integrating resistive memories that are well suited to emulate artificial plastic synapses at ultra-low power[41–45]. Given their unique features, SNNs thus constitute promising candidates for future very-low power and fully unsupervised neural signal processing embedded within cortical implants. However, SNN architectures are typically dependent on the application for which they are dedicated and are not versatile to answer the need of a wide range of different applications. Frugal SNN-based algorithms thus remain to be developed to enable fully unsupervised and very-low-power and online-compatible neural pattern detection and classification from large-scale multivariate neural recordings.

A number of previous studies have shown that STDP-based SNN architectures could be used for pattern or object recognition within static images in either supervised[46–48], semi-supervised[49], or fully unsupervised[50–53] ways. Other architectures have been developed for supervised learning of patterns within a temporal signal such as speech[54]. Such a paradigm has further been extended to the supervised recognition or decoding of dynamic patterns within time-varying multivariate data such as EMG[55], olfactory signals[56], respiratory[57] signals, tactile braille reading[58], EEG[59], or intracortical data[60,61]. Toward unsupervised learning and classification of multivariate patterns, hybrid approaches have been developed consisting of a self-organizing SNN trained in an unsupervised way to represent an audio input followed by a supervised step to classify each representation into a sound class[62,63]. These networks classically have a feed-forward structure with several fully-connected layer, resulting in large numbers of parameters[58], which can be reduced to some extend using convolutional layers[47,62]. However, very frugal networks with very few elements remain to be developed. Moreover, the fully unsupervised extraction and classification of time-varying patterns using SNNs is a problem that has been much less investigated previously.

In a previous study, we proposed an attention-based SNN to extract and automatically classify action potential shapes from single-channel extracellular neural signals[64]. A next challenge is the extraction of multivariate temporal patterns in a fully unsupervised way[65]. In this respect, previous works have tackled the specific case of time-varying visual scenes. In particular, efficient fully unsupervised learning of spatio-temporal patterns corresponding to moving objects has been demonstrated using SNNs[66,67]. When searching for spatio-temporal patterns, the temporal information needs to be somehow captured by the network. In one of these studies, data from AER cameras could be automatically processed to count cars passing in each line of a highway[66]. In this case, spatiotemporal patterns typically had similar dynamics, such as their duration in time and how fast they moved across pixels. Moreover, this strategy was based on the order of spikes within the input, so that the output neurons eventually fire after only a few spikes of the patterns are emitted[27,28], not waiting for the whole patterns to end. As a consequence, such approach typically fails when different spatiotemporal patterns are nested, for example one pattern being exactly the beginning (along either space or time or both) of another longer one, so that both only differ by their endings. In another study[67], feedforward connections between neurons with different membrane dynamics were used to achieve memory of different time-scales in order to learn temporal patterns with different dynamics. An alternative strategy we employed in a previous study to automatically classify temporal patterns corresponding to different action potential shapes in an extracellular neural signal was to use several synapses between two neurons with different delays[64]. The drawback of these approaches is the multiplication of the number of neurons and/or synapses. An alternative could be to segment the flow of temporal data into fixed-size frames and process each frame like an image. This however prevents fully online processing of the temporal data without any prior on the length of the patterns to search for. To overcome these limitations, there is thus a need for very frugal architectures able to automatically configure to recognize multivariate temporal patterns in a fully unsupervised way and compatible for online processing of data streams.

Toward this goal, we propose here an architecture that highly contrasts with classical ones based on Leaky-Integrate-and-Fire (LIF) neurons[68]. Among the wide range of existing models of spiking neurons[68–72] we used a variant of LTS neurons[72], whose dynamics automatically adapt to the temporal durations of input patterns without the need to multiply the number of synapses. A single layer of such neurons is connected to input spike trains by synaptic weights and the network learns through biological learning rules like STDP and Intrinsic Plasticity (IP) in a fully unsupervised manner. We show that, with only a handful number of neurons, this strategy is efficient to recognize highly overlapping multivariate temporal patterns, first on simulated data, and then on Mel cepstral representations of speech sounds and finally on multichannel neural data. These results thus pave the way toward highly frugal SNNs for unsupervised learning of complex multivariate temporal patterns.

## Results

Ideally, an SNN developed for unsupervised pattern identification and recognition should be able to process incoming data fed sequentially, and emit one output spike each time a specific pattern occurs in the data, and each spike needs to be emitted by a different neuron for each different pattern to allow direct inference without further supervised step. The initial step of this procedure is to encode the continuous input data into spike trains to enable SNNs to leverage the inherent event-driven nature of neuronal computation and capture temporal dependencies within the data. To this end, the original data was quantized using a column of sensory receptive fields that generated spikes when the signal fell within the fields (Figure 1a), a strategy we previously employed to encode extracellular signals for spike sorting using SNNs[64]. Five spikes were generated for each input value in order to increase robustness of the encoding with respect to small fluctuations of the input. Using this approach, the resulting spike trains directly reflected the shape of the original data (Figure 1b). Original audio data was decomposed into 24 continuous Mel Cepstral signals (Figure 1c, see Methods for details). Similarly, the multiunit activity of the neural data was binned and smoothed for each electrode, resulting in 30 continuous signals (Figure 1d). In both cases, these input data were normalized between 0 and 1 and encoded using 24 receptive fields (Figure 1e,f).

This initial encoding resulted in spike trains that encoded both background noise and relevant patterns. In order to mitigate the influence of noise on the STDP learning process, a short-term (STP) plasticity rule was introduced for each input spike train, which is a mechanism that weakens input synaptic weights all the more that the presynaptic activity is high (see Methods). With this strategy, only those spike trains encoding the peaks and throughs in the original data were retained (Figure 1g,h). These final encoding spike trains were then considered as the input spike trains passed into the network to be processed.

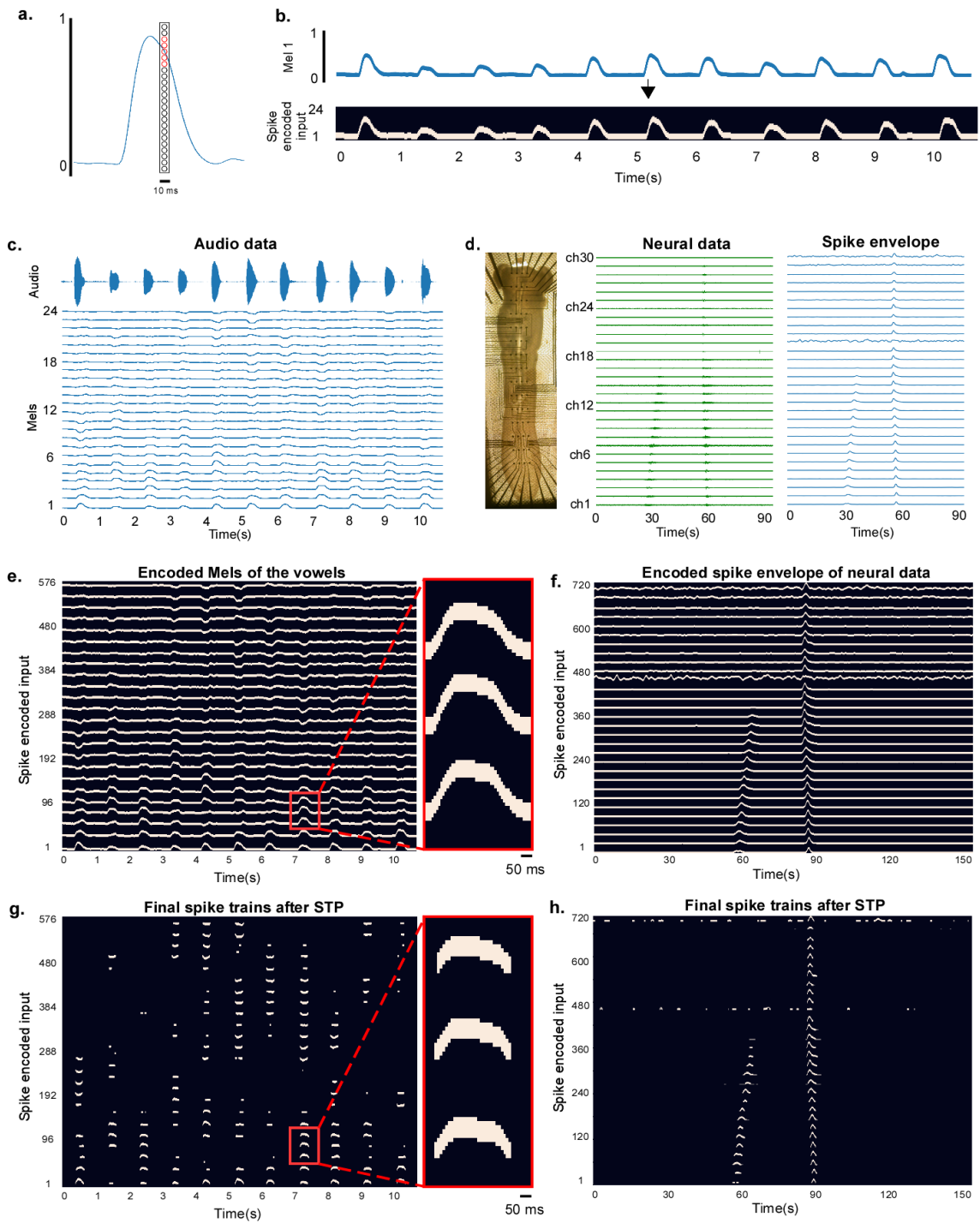

**Figure 1 | Spike encoding pipeline. a** Each signal was normalized and converted into 24 spike trains using quantization receptive fields. At each timestep, depending on the value of the signal, a spike was generated by one of 20 receptive fields equally spanning the 0-1 range of input values. Two additional spikes were generated both above and below the central spike making a total of five spikes per timestep (receptive fields in red) and 24 spike trains per signal. **b** Example of the normalized Mel 1 corresponding to eleven French vowels and encoded into a 24 spike trains. **c** Audio data decomposed into 24 Mel cepstral coefficients. **d** Multiunit neural data (middle) from embryonic mouse hindbrain spinal cord on MEA (left) was binned and smoothed to extract spike envelopes on each channel (right). **e,f** Initial encoding spike trains for audio and spike envelope of neural data, respectively. **g,h** Final encoding spike trains after STP for audio and spike envelope of neural data, respectively. STP eliminated spikes corresponding to noise and only retained spikes indicative of a pattern. Some residual spikes corresponding to noise can be seen across some neural data channels as these channels were very noisy.

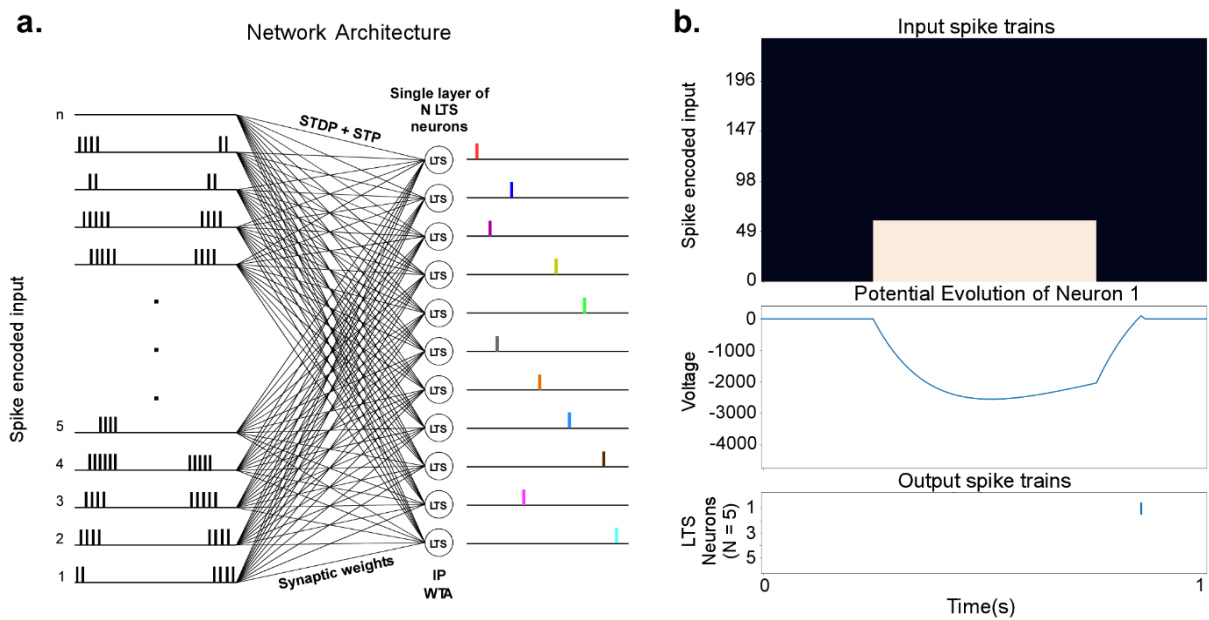

**Figure 2 | Network architecture. a** Inputs encoded as spike trains were passed into the network consisting of only a handful of LTS neurons. Learning happened through STDP and IP rules that enabled the LTS neurons to modulate their input synaptic weights and thresholds, respectively. A Winner-Take-All (WTA) mechanism in the LTS layer chooses the neuron with the steepest rebound among the neurons that have crossed their respective thresholds to produce an output spike (see Methods) thereby ensuring that at most only one neuron emits a spike at any given timestep. **b** Illustration of the working mechanism of an LTS neuron with a simple case of 250 input spike trains (top: black = 0 or 'no spike'; white = 1 or 'spike') to a single LTS neuron. The voltage of the LTS neuron (middle) is inhibited by the incoming example stimulus for a duration that is determined by the time constant of the neuron. Upon the end of incoming stimulus, the neuron generates a rebound potential and emits a spike (bottom) when it crosses a threshold. With every postsynaptic spike, the voltages of all neurons were reset to 0. This behavior is ideal as the neurons waited for the pattern to end and produced only a single output spike per pattern.

Shown in Figure 2a is the final network architecture, which consisted in a single layer of LTS neurons that were connected to the input spike trains by negative synaptic weights (between -1 and 0) initialized randomly according to a uniform distribution. The LTS neurons processed the spikes from the input spike trains (presynaptic spikes) and sporadically produced output spikes (postsynaptic spikes) according to the dynamics of their membrane potentials. A Winner-Take-All (WTA) mechanism implemented across the LTS neurons ensured that there was at most only one neuron that emitted a spike at any given timestep. Learning in the network happened through biological learning rules such as STDP and Intrinsic Plasticity (IP). For each postsynaptic spike emitted by a certain neuron in the network, the STDP rule strengthened the synaptic weights between this neuron and all the spike trains that had presynaptic spike within a certain coincidence time window chosen based on the maximum length of patterns to be searched for in the data. Another lateral STDP rule governing lateral inhibition weakened the synaptic weights between all the other neurons and the same spike trains thereby inhibiting other neurons from learning the same pattern (see Methods). The lateral STDP update rule was much weaker than the principal STDP rule, considering that patterns may share common spiking activity. Additionally, for each postsynaptic spike output by a certain neuron, the IP rule adapted the threshold of the neuron based on the size of the pattern learnt giving the network the ability to differentiate between patterns that are one inside the other. Figure 2b demonstrates the evolution of the membrane potential of an LTS neuron in the presence of multiple input spike trains. Each input spike inhibited the potential of the LTS neuron for a duration that was determined by the neuron's membrane time constant. Once the stimulus ended, the neuron generated, by nature of the way they were modeled, a rebound. At the network level, the neuron with the highest inhibition produced the steepest rebound according to which it was the chosen neuron to emit an output spike (see Methods). Therefore, the network took in multiple spike trains representative of the original data as input and processed these spikes to produce sporadic output spike trains that corresponded to patterns in the input data, all while learning through biological learning rules in a fully unsupervised manner.

In order to establish a baseline for the network's classification capabilities, the network was initially tested with artificial patterns that mimicked spike trains obtained from encoding spectrograms (Figure 3). In this approach, artificial patterns mimicking unique frequency characteristics across 240 spike trains were repeatedly shown to the network to learn. Starting with basic patterns, and increasing the complexity of the patterns iteratively helped us better troubleshoot the classification performance of the network. The most basic example of artificial patterns included four non-overlapping patterns (Figure 3a). The network was trained on 50 repetitions (= epochs) of these four patterns and it can be seen that the network was able to identify these four unique patterns and each pattern was learned by a unique LTS neuron right from the beginning (Figure 3b). Once all the input spikes are passed through the network, the output spike trains produced by the network were matched with the truth spike trains to obtain truth-output pairs to compute an evolving f-score along the learning process. It can be seen that for these simplest patterns, the network had a perfect f-score of 1 since the beginning. This can be attributed to the random initialization of the input weights to the LTS neurons and the absence of commonality in spike trains between patterns.

In both neural and vocal datasets, it is not uncommon to encounter patterns that are embedded within each other. Therefore, we tested the network with another example of four patterns (also repeated 50 times), where two were subsets of two other ones in terms of frequency characteristics (Figure 3c). It can be seen that at the beginning of the training, two LTS neurons each learned two input patterns where one is inside the other (between epochs 7 and 38, see Figure 3d). However, as learning continued, IP helped the neurons to adapt their threshold thereby preventing them to spike for the smaller patterns, which could then be learnt by two other neurons (the f-score eventually evolved to 1 as learning progresses).

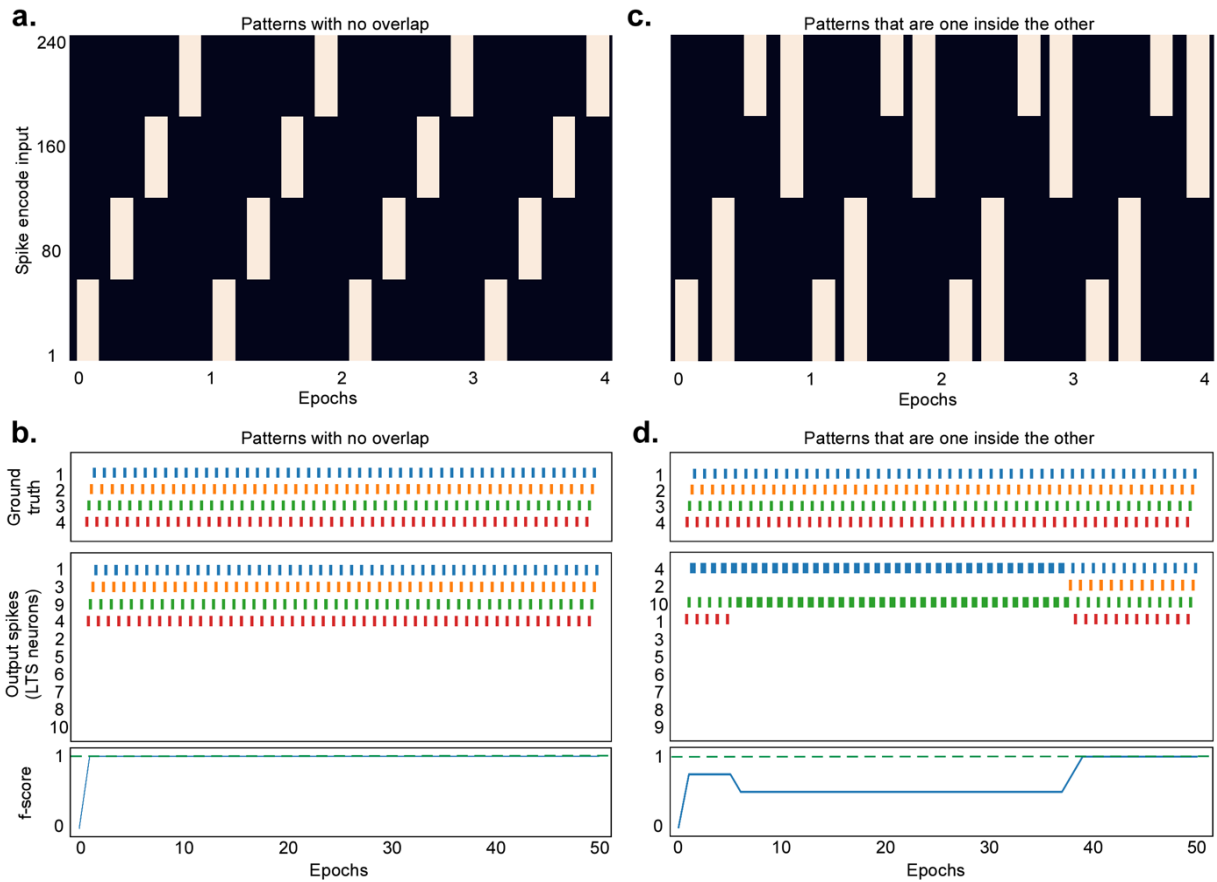

**Figure 3 | Classification performance on artificial patterns. a** Case of four patterns that do not overlap in terms of frequency characteristics across 240 spike trains. **b** Four of the ten LTS neurons in the network were able to identify and learn each of the patterns. The f-score was 1 throughout the training. **c** Case of patterns with two being inside two others. Intrinsic Plasticity helped neurons adapt their thresholds in order to be able to differentiate between the small and big patterns. The f-score reached 1 as learning progresses.

After having established a baseline for the network's classification capabilities with the artificial patterns, we then tested the network on real speech data consisting of eleven French vowels repeated 50 times by a native French speaker (Figure 4a, top). The network was trained on 20 epochs of encoded spike trains from this data with eleven LTS neurons and at the end of the training, the output spike trains and the truth spike trains were matched to obtain the best truth-output pairs. After a learning period on as little as 20 epochs, the output spikes produced by the network became coherent with respect to the ground truth sequence of produced vowels (Figure 4a, bottom). The f-score obtained on the classification performance of the network on the final epoch was 0.92 (see also the corresponding confusion matrix shown in Figure 4b). The performance then remained stable after the 20$^{th}$ epoch if the data was continued to be fed to the network. At the end of training, each LTS neuron had learned a unique vowel as reflected by the final weights of the neuron that were strong for the Mels corresponding to this vowel (Figure 4c). Figure 4d illustrates the evolution of the synaptic weights of neuron 1 (see cyan spikes in the bottom raster of Figure 4a and first column of Figure 4c), evolving from random initial values to final values.

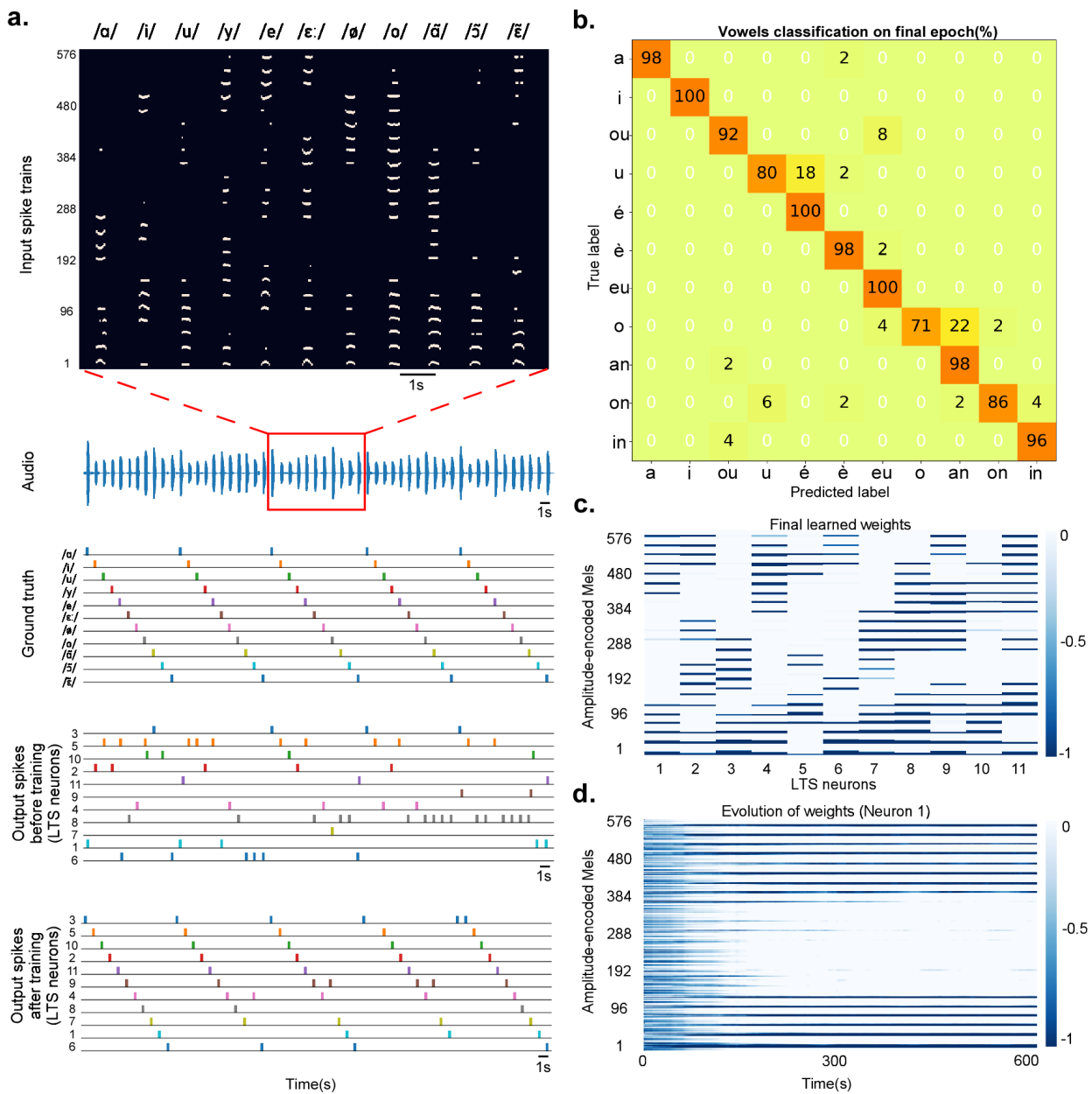

**Figure 4 | Unsupervised recognition of vowels. a** Eleven French vowels were repeated several times and the encoding spike trains corresponding to one repetition are visualized. At the beginning of training, the output spikes produced by the network were random with respect to ground truth, whereas, at the end of training, each pattern was learned by a unique LTS neuron. **b** Confusion matrix computed on the final epoch of training. **c** Final weights of the eleven LTS neurons after training. **d** Evolution of the weights of neuron 1 through time.

In a last step, we tested whether the network could also automatically identify spatiotemporal patterns in multielectrode array neural data. Neural activity was recorded in a whole embryonic hindbrain and spinal cord preparation, which was previously shown to exhibit rhythmic propagating waves of activity[73,74]. The multiunit spiking activity envelope of all these channels were encoded through the encoding mechanism as illustrated in Figure 1h. This preparation exhibited a short and a long spiking patterns that were repeating in time. The short pattern had spiking activity propagating caudo-rostrally across the lower channels covering the lower thoracic and lumbar/sacral region of the spinal cord, whereas the long pattern had spiking activity propagating rostro-caudally across all channels. The network was trained on 10 epochs of the encoded spike trains with five output neurons. After 5 epochs, two unique output neurons had learnt the short and long patterns (Figure 5a), producing consistent and coherent output spikes with respect to the ground truth (fscore=1). After completion of the 10 epochs, the final weights of the LTS neurons (Figure 5b) confirmed the learning of the short pattern by neuron 3 and the long pattern by neuron 4, with, in both cases, strong weights reflecting the corresponding patterns.

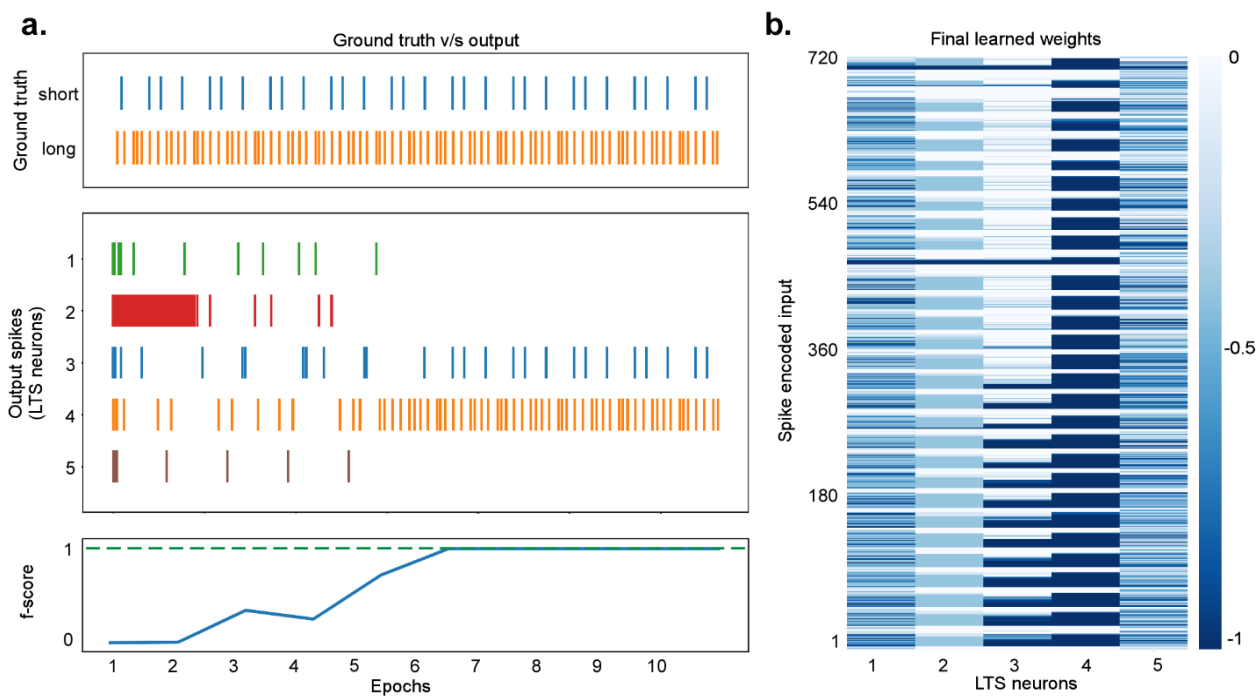

**Figure 5 | Unsupervised recognition of rhythmic activity patterns in neural data. a** Neural data consisted of 3 short and 9 long spiking patterns over a period of 1h. The network was trained on 10 repetitions of this recording. It can be seen that as training progresses, output neuron 3 and 4 had learnt to identify the short and long patterns, respectively. The f-score reached 1 after 5 epochs. **b** Final weights for each LTS neuron confirming the learning where neurons 3 and 4 had strong weights for the inputs corresponding to short and long patterns, respectively.

## Discussion

The primary objective of this study was to explore the capabilities of a novel SNN architecture for unsupervised pattern classification in multichannel temporal data. In this work, we presented a very frugal single-layer SNN capable of learning patterns in continuous streams of data through unsupervised biological learning rules. The encoding method used ensured that the input spike trains only contained spikes during the presence of a pattern. We tested the network's classification capabilities on three types of data, namely simulated artificial patterns, audio data containing eleven different vowels, and neural data across 30 channels recorded from a spinal cord. The artificial data helped us establish a baseline for the network's classification capabilities and confirmed learning through biological learning rules. The analysis of vowel classification revealed that the network maintained robustness across multiple instances of identical vowels, despite possible variability in their pronunciation. Classification on the spike envelopes of neural data was a proof of concept of fully unsupervised pattern recognition within multielectrode array data. Furthermore, our choice of using LTS neurons as the spiking neuron model enabled us to have a single layer of only a few neurons based on the fact that the neurons wait for the pattern to end and produce one postsynaptic spike after the end of the pattern. As a result, one neuron produced one output spike per pattern for easy inference. We showed that the combination of the LTS neurons and several biological plasticity rules resulted in each pattern requiring just one neuron to be learned. Existing research in the domain of unsupervised or even supervised classification with SNNs often involve multiple convolutional layers of LIF neurons to extract features from input data, thus typically involving several hundred thousand parameters that need to be learned[66,67], precluding highly energy-efficient neuromorphic pattern learning. Sometimes, they also involve an additional post-hoc classifier at the readout layer to classify the aforementioned features[62,63]. The learning paradigm is also different as training is usually done non sequentially by splitting the continuous data into frames of fixed duration, as opposed to the continuous learning in our case.

A significant feature of the proposed algorithm and SNN architecture, is its ability to discriminate between highly overlapping patterns, and in particular between patterns where one is completely included into another one. This was possible thanks to the use of an IP rule that allowed the LTS neurons to adapt their thresholds based on the size of the patterns. The network is further dependent upon only a few critical parameters. The first one is the LTS neuron time constant. It determines the inertia of the neuron before generating a spiking rebound, and should thus be chosen according to the inter-spike intervals within the input spike trains and the expected interval between patterns. The second important parameter is the STDP lookback window, which determines the maximum duration of the patterns that are being searched for. If too short, it might prevent learning long patterns, and if too long and exceeding the minimum interval between patterns, it might aggregate patterns.

Our results on neural data are a first proof of concept of fully unsupervised and online-compatible recognition of spatiotemporal patterns within multielectrode array data. Neural data are constituted of two types of signals, spiking activity reflecting the emission of action potentials, and local field potentials (LFPs) reflecting all other type of transmembrane neural currents and in particular synaptic activity. We thus deliberately chose to address the problem of pattern extraction from the envelope of spiking activity collected over multiple channels. This makes the approach directly applicable to low frequency LFP signals and to envision algorithms based on frugal SNNs for very-low-power neural feature extraction embedded in future active neural implants.

# Methods

## Network Architecture

### LTS neurons

As illustrated in Figure 2a, the network consisted of a single layer of a few LTS neurons that were connected to input spike trains through negative synaptic weights initialized randomly according to a uniform distribution and further clipped between [-1,0] at all times. LTS neurons, that are a type of Integrate-and-Fire (IF) neurons, have the property to be inhibited during the presence of a stimulus and generating a rebound after the end of the stimulus (Figure 2b). Therefore, as the spike trains are passed through the network, incoming currents (spikes x weights) due to the presence of a pattern hyperpolarized all the LTS neurons. Once the incoming currents stopped due to the end of the pattern, the LTS neurons generated a potential rebound. A Winner-Take-All (WTA) mechanism chose among the neurons that have crossed their respective thresholds, the one with the steepest rebound to generate a postsynaptic spike. The LTS neurons were modeled by the following equations:

$$\tau_m \frac{dV}{dt} = -V + q + gI_{stim}$$

$$\frac{\tau_m}{\varepsilon} \frac{dq}{dt} = -q + f(V)$$

$$with\ f(V) = \begin{cases} \alpha_n V\ if\ V < 0 \\ \alpha_p\ if\ V \geq 0 \end{cases}$$

where $V$ is the LTS neuron potential, $q$ is an adaptation variable that triggers the rebound after inhibition, $\tau_m$ is the membrane time constant that is chosen depending on the type of data, $\varepsilon$ is a constant that makes $q$ vary slower than $V$, $I_{stim}$ is the stimulus current (spikes x weights) of the timestep and $g$ is a constant. Whenever the network produces a postsynaptic spike, both $V$ and $q$ were reset to 0 for all neurons. The following table illustrates the LTS neurons parameters used for the different types of data the network was tested on.

| Parameter | Artificial Patterns | Vowel data | Neural data |
|---|---|---|---|
| $\varepsilon$ | 0.001 | 0.01 | 0.001 |
| $\alpha_n$ | -200 | -200 | -200 |
| $\alpha_p$ | -10 | -10 | -10 |
| $\tau_m$ | 15 | 15 | 50 |
| $g$ | 100 | 100 | 100 |

Table 1 | Core parameters of the LTS neurons for the different types of data tested.

The membrane time constant $\tau_m$ was chosen according to the size of the patterns expected in the input data. The artificial patterns and vowels data contained patterns that lasted for about ~500 ms on average. Neural data, on the other hand contained patterns that lasted several seconds. The parameter $\varepsilon$ was chosen according to the inter-pattern interval in the data.

**Plasticity rules**

Learning took place whenever a postsynaptic spike was output by the network. At the occurrence of every postsynaptic spike, the following plasticity rules enabled the network to learn:

**Classical STDP** strengthened the synapses connecting the neuron that generated a postsynaptic spike and the input spike trains that exhibited spiking activity within a certain pre-time window, thereby implementing Long-Term Potentiation (LTP). It also weakened the synapses connecting the same post-synaptic neuron and the input spike trains that did not exhibit any spiking activity within the pre-time window, thereby implementing Long-Term Depression (LTD). We chose to implement a simple version of this rule, which is define as follows:

$$\Delta w_{ij} = \begin{cases} w_{LTP}, & \text{if } \exists t_i \in S_i \text{ such that } t_j - T_{STDP} < t_i \leq t_j \\ w_{LTD}, & \text{if } \nexists t_i \in S_i \text{ satisfying } t_j - T_{STDP} < t_i \leq t_j \end{cases}$$

where $w_{ij}$ is the synapse connecting input spike train $i$ and the LTS neuron $j$ that spiked, $t_i$ is the time of occurrence of the presynaptic spike and $t_j$ is the time of occurrence of the postsynaptic spike, $S_i$ is the set of presynaptic spike times for the input spike train $i$, $T_{STDP}$ is the duration of the window preceding $t_j$ that determines the relevant temporal context for STDP, $w_{LTP}$ = -0.1 as we use negative weights, and $w_{LTD}$ = 0.06. $T_{STDP}$ was set to 500 ms for the artificial patterns and the vowel data and 8 seconds for the neural data.

**Lateral STDP**, another STDP rule was used to govern lateral inhibition between LTS neurons. It weakened the synapses connecting all neurons other than the postsynaptic neuron that spiked, and the input spike trains that exhibited spiking activity within the same pre-time window. This was to prevent multiple neurons from learning the same pattern. However, this update rule was much weaker than the classical STDP rule, keeping in mind that patterns might share common spiking activity. For the postsynaptic neuron $j$ that spiked, the lateral STDP rule is defined as follows:

$$\Delta w_{ij} = w_{potentiation}, \text{ if } \exists t_i \in S_i \text{ such that } t_j - T_{STDP} < t_i \leq t_j$$

For all other postsynaptic neurons $k \neq j$, the lateral STDP rule is defined as:

$$\Delta w_{ik} = w_{inhibition}, \forall k \in N, k \neq j, \text{ if } \exists t_i \in S_i \text{ such that } t_j - T_{STDP} < t_i \leq t_j$$

where $w_{ik}$ is the synapse connecting input spike train $i$ and each non-spiking postsynaptic neuron $k$, $N$ is the set of all postsynaptic neurons, $w_{inhibition}$ = 0.0002 and $w_{potentiation}$ = -0.001. This formulation drove the network towards a more selective and refined connectivity pattern based on the temporal spiking relationships.

**Intrinsic Plasticity**, unlike STDP, is a form of plasticity implemented on the neurons and not the synapses connecting the spike trains and the neurons. It helped neurons adapt their thresholds based on the size of the pattern learned. The thresholds of all output neurons were initialized at a low value, to promote learning at the beginning of training and as training progressed, each neuron increased its threshold $Th$ according to the size of the pattern learnt and reached an equilibrium threshold indicative of the size of the pattern learnt. Every time a postsynaptic neuron emitted a spike, its threshold $Th$ was decreased by $\Delta Th_{post} = F^{\Delta Thpost} * Th$. For each pre-synaptic spike received within a coincidence time window before the post-synaptic spike, the threshold was increased by a value that was obtained by multiplying $\Delta Th_{pair}$ by the synaptic weight. The thresholds of all neurons were initialized at 20 and then were clipped between [20,3500] at all times.

| Parameter | Artificial Patterns | Vowel data | Neural data |
|---|---|---|---|
| $F^{\Delta Thpost}$ | 0.01 | 0.01 | 0.5 |
| $\Delta Th_{pair}$ | 0.1 * $F^{\Delta Thpost}$ | 0.6 * $F^{\Delta Thpost}$ | 0.1 * $F^{\Delta Thpost}$ |

**Table 2** | Threshold update parameters of the LTS neurons for the different types of data tested.

## Encoding

The process of transforming multichannel data into spike trains is a pivotal step in training SNNs for learning tasks. The effectiveness of this encoding directly influences the network's ability to classify and interpret data. The encoding method determines how well the temporal and spatial dynamics of the data are captured and represented as spikes. Here, we encoded each channel of the data into a collection of spike trains while retaining the original geometry of the data. As shown in Figure 1a, each channel is normalized and discretized into 20 receptive fields. The continuous signal, ranging from 0 to 1, was divided into 20 equal intervals representing the sensitivity of each receptive field. At each timestep, depending on the value of the signal, a spike was encoded by the field corresponding to the signal value. Two additional spikes were encoded both above and below the central spike making a total of five spikes per timestep. There was therefore 24 spike trains representing each channel of the data. The artificial patterns did not have an encoding step as they already represented the final spike trains ready to be passed into the network.

### Short-Term Plasticity

To ensure that learning by LTS neurons is not driven by the background noise of all channels, we implemented a mechanism called Short-Term-Plasticity (STP) that quickly suppressed all the spike trains that corresponded to noise/silence. After the unwanted spike trains were suppressed, the retained spike trains were the ones that encoded rich vowel information. To implement STP, we assigned a weight $w_{STP}$ to every spike train. This weight, which was initialized to 1 for all spike trains, is a probability of the spike train to encode a signal. The input spike trains were subjected to STP before training and as they are processed through time, the weights of the spike trains encoding noise are quickly decreased. Once the weight of any spike train fell below 0.75, we stopped STP and mapped the weights of all spike trains below a certain threshold to 0 and the others to 1. This threshold was 0.92 for the vowels and 1 for the neural data. Furthermore, for each group of 24 spike trains corresponding to a certain signal, we checked if at least 60% of the spikes were retained after STP and if not, the remaining spikes were also mapped to 0 in order to clean up residual spikes potentially corresponding to noise. STP is governed by the following equations:

$$\frac{dw_{STP}}{dt} = \frac{1}{\tau_{stp}}(1 - w_{STP})$$

$$\frac{dw_{STP}}{dt} = \frac{1}{\tau_{stp}}(1 - w_{STP}) - w_{STP} * f_d$$

where $\tau_{stp}$ = 2000 ms is the STP time constant and $f_d$ = 0.003 is the depression factor. The first of these two equations is the weight update rule for spike trains that do not have spikes in the timestep and the second equation is the weight update rule for spike trains that have spikes in the timestep. Post STP, only the spikes corresponding to the spike trains encoding relevant data beyond noise were retained (see Figure 1g,h).

**Vowel data**

The vowels were recorded with a microphone (SHURE Beta 58 A) and Audacity software at a sampling rate of 44.1 kHz. A native French male was asked to repeat eleven French vowels 50 times. The recorded audio was subjected to a frequency transform using the SPTK library to obtain 25 Mel Cepstral coefficients. The first Mel reflecting mostly the amplitude of the sound, and thus being not specific to which vowel was pronounced, was discarded. The other 24 Mels were normalized between 0 and 1, smoothed, quantized and encoded as spike trains (see Figure 1c,e,g) into an array of binary values. Prior to encoding, we chose to smoothen the Mels with a sliding 2$^{nd}$ order Butterworth filter below 5 Hz to make the network more robust to different occurrences of the same pattern. Unlike the artificial patterns, which had spikes only during the duration of each pattern and no spikes before or after the pattern, the vowels' spike trains had spikes corresponding to noise/silence across all Mels. Therefore, the encoded spike trains were first subjected to STP, to eliminate spikes corresponding to noise and to only retain spikes corresponding to peaks and throughs. These spike trains were then passed into the network as input.

**Neural data**

Neural data was reused from a previously published study[74]. They corresponded to rhythmic activity waves propagating across a whole embryonic OF1 mouse hindbrain-spinal cord preparation at stage E13 laid down on a 60-channel microelectrode array (Ayanda Biosystems, Lausanne Switzerland) arranged as 4 columns of 15 microelectrodes (Figure 1d, left). The detailed procedure to acquire these data has been detailed previously[74] and was in accordance with protocols approved by the European Community Council and conformed to National Institutes of Health Guidelines for care and use of laboratory animals. In short, after dissection and meninges removed, the neural tissue was maintained on the electrode array with a custom net and continuously superfused with aCSF (in mM: 113 NaCl, 4.5 KCl, 2 $CaCl_2 2H_2O$, 1 $MgCl_2 6H_2O$, 25 $NaHCO_3$, 1 $NaH_2PO_4 H_2O$, and 11 D-glucose) at a rate of 2 ml/min. Neural data were acquired at 10kHz using a MEA1060 amplifier from Multi Channel Systems (MCS), with x1200 gain and 1–3000 Hz bandpass filters, connected to two synchronized Power 1401 acquisition systems (Cambridge Electronic Design LTD, Cambridge, UK). Each channel was then bandpass filtered between 200 Hz and 2kHz to retain high-frequency components (Figure 1d, middle). Once filtered, we extracted multiunit activity by computing the mean and standard deviation of each channel and considered as spikes those datapoints that were at least 3 standard deviations above or below the mean. Each channel was then downsampled by a binning factor of 100 where each bin was replaced by the total number of spikes in the bin. Finally, a Gaussian kernel (n = 501, σ = 51 time bins) was convolved to each channel to obtain smoothed spike envelopes of the original neural data (Figure 1d, right). These spike envelopes were then normalized between 0 and 1 and encoded as spike trains and subjected to STP in a manner similar to the vowels. The final spike trains obtained after STP were then passed into the network for learning.

**Inference and Evaluation**

To assess the classification performance of the network, we first matched the truth spike trains and the output spike trains to get truth-output pairs. To perform this matching, we convolved all the truth spike trains and output spike trains with a Gaussian kernel (n = 31, σ = 3 time steps) and then computed the cross-correlation between each of the truth spike trains and the output spike trains. For each truth spike train, we chose the output spike train with the highest cross-correlation as the corresponding output. For each pair, the f-score was computed as:

$$F_{ij} = \frac{2 * H_{ij}}{T_i + O_j}$$

where $T_i$ was the number of spikes of the $i$th truth spike train, $O_j$ was the number of spikes emitted by the $j$th output neuron and $H_{ij}$ was the number of output spikes coinciding with a truth spike within a coincidence window. The coincidence window was 400 ms for the artificial patterns and vowel data and 2.5 seconds in the case of the neural data. These values corresponded to the time needed by a LTS neuron to generate its rebound and cross its threshold. We also computed a global f-score across all truth neurons and all output neurons as:

$$F = \frac{2 * H}{T + O}$$

where $T$ was the total number of truth spikes, $O$ was the total number of output spikes and $H$ was the total number of hits. In the case of the vowels, a confusion matrix was also computed to evaluate the classification performance of the model.

# Acknowledgements

This work was supported by the French Research National Agency through the ANR-20-CE45-0005 BrainNet project.

# Data and code availability

The code and data used in this study will be made available upon acceptance of the manuscript to reproduce the results.